\documentclass[twoside,twocolumn]{article}

\usepackage{blindtext} 

\usepackage[sc]{mathpazo} 
\usepackage[T1]{fontenc} 
\usepackage{microtype} 

\usepackage[english]{babel} 

\usepackage[hmarginratio=1:1,top=32mm,columnsep=20pt]{geometry} 
\usepackage[hang, small,labelfont=bf,up,textfont=it,up]{caption} 
\usepackage{booktabs} 

\usepackage{lettrine} 

\usepackage{enumitem} 
\setlist[itemize]{noitemsep} 

\usepackage{abstract} 

\usepackage{titlesec} 
\renewcommand\thesection{\Roman{section}} 
\renewcommand\thesubsection{\roman{subsection}} 
\titleformat{\section}[block]{\large\scshape\centering}{\thesection.}{1em}{} 
\titleformat{\subsection}[block]{\large}{\thesubsection.}{1em}{} 

\usepackage{fancyhdr} 
\usepackage{titling} 

\usepackage[colorlinks=true,linkcolor=blue, urlcolor=blue, citecolor=blue]{hyperref} 

\usepackage{graphicx}

\usepackage{caption} 
\let\oldbibliography\thebibliography
\renewcommand{\thebibliography}[1]{%
	\oldbibliography{#1}%
	\setlength{\itemsep}{5pt}%
}

\pretitle{\begin{center}\Huge\bfseries} 
\posttitle{\end{center}} 
\title{On Solving the 2-Dimensional Greedy Shooter Problem for UAVs} 
\author{%
\textsc{Loren Anderson}\thanks{This material is based upon work supported by the National Science Foundation Graduate Research Fellowship Program under Grant No. 00074041. Any opinions, findings, and conclusions or recommendations expressed in this material are those of the author(s) and do not necessarily reflect the views of the National Science Foundation.}\\[1ex] 
\normalsize University of Minnesota Twin Cities \\ 
\normalsize \href{mailto:and05097@umn.edu}{and05097@umn.edu} 
\and 
\textsc{Sahitya Senapathy}\\[1ex] 
\normalsize St. Mark's School of Texas \\ 
\normalsize \href{mailto:sahitya.senapathy@gmail.com}{sahitya.senapathy@gmail.com} 
}
\date{\today} 
\renewcommand{\maketitlehookd}{%
\begin{abstract}
\noindent Unmanned Aerial Vehicles (UAVs), autonomously-guided aircraft, are widely used for tasks involving surveillance and reconnaissance. A version of the pursuit-evasion problems centered around UAVs and its variants has been extensively studied in recent years due to numerous breakthroughs in AI. We present an approach to UAV pursuit-evasion in a 2D aerial-engagement environment using reinforcement learning (RL), a machine learning paradigm concerned with goal-oriented algorithms. In this work, a UAV wielding the greedy shooter strategy engages with a UAV trained using deep Q-learning techniques. Simulated results show that the latter UAV wins every engagement in which the UAVs are suffciently separated during initialization. This approach highlights an exhaustive and robust application of reinforcement learning to pursuit-evasion that provides insight into effective strategies for UAV flight and interaction.
\end{abstract}
}

\begin{document}

\maketitle


\section{Introduction}

Unmanned Aerial Vehicles (UAVs), autonomously-guided aircraft, are widely used for tasks involving surveillance and reconnaissance. Due to their low cost and versatility, UAVs are ideal for artificial intelligence applications in physical domains. Conveniently, the dynamics of UAVs are easily replicated in simulated environments, yielding rapid data collection. This is precisely the situation in which a popular machine learning technique, reinforcement learning (RL), flourishes: the speed of simulated environments supports the learning of a goal-directed policy through direct experience. Some applications of RL to UAVs include tracking \cite{2019Akhloufi, 2019Arola, 2019Bonnet, 2018Ma, 2019Wang}, navigation \cite{2013Faust, 2010Ferrari, 2016Imanberdiyev, 2019Koch, 2017Wang, 2015Zhang}, and pursuit-evasion \cite{2019Akhloufi, 2019Arola, 2016Camci, 2012Li, 2018Pippin}. Inspired by recent successes of applying RL techniques to adversarial games, we focus on the 2D pursuit-evasion game between two UAVs here. Notable triumphs have occurred in the games of Go \cite{2016Silver, 2017Silver}, Atari \cite{2015Mnih}, and StarCraft II \cite{2019Vinyals}, in which the trained RL agents achieved superhuman performance. These results are particularly enticing, as the agents illuminated novel strategies, some without incorporating prior human knowledge.

In this paper, we follow the framework of \cite{2018Pippin} in which a UAV trained with RL techniques engages against another UAV, dubbed the \textit{greedy shooter} (GS), that acts to position itself facing the former UAV. Section \ref{prob} formalizes the problem in terms of Markov decision processes and describes the environment. In Section \ref{methods}, we explain our approach and deep Q-learning implementation. Performance is discussed in Section \ref{result} where we provide a brief overview of the RL agent's learned strategies. Section \ref{con} contains concluding remarks, and the appendix in Section \ref{app} showcases sample trajectories.


\section{Problem Description}

\label{prob}

This is a single-agent RL problem in 2D space with two agents, the RL and GS UAVs. Hereafter, we refer to these UAVs as the RL and GS agents. The goal of the RL agent is to catch the GS agent within its \textit{targeting zone}, and vice versa, at the end of each discrete time step. The RL agent trains and continually updates its strategy, known as its policy; in contrast, the GS agent has a fixed policy in which it turns to align its heading in the direction of the RL agent. This policy, known as pure pursuit \cite{1985Shaw}, is well-known to be reasonably effective in our proposed environment.

We model the problem as a fully-observable Markov decision process (MDP) with the 5-tuple $(S, A, P, R, \gamma)$. The state, $S$, is a 6-dimensional vector containing the positions and headings (forward directions) of each agent. The action space, $A$, of each agent details its possible speeds and turn angles. At the beginning of each time step, the agent supplies an action to the environment from the permissible pairs of speeds and turn angles in its action space. The given angle and speed instantaneously update the agent's current heading and speed, respectively, remaining fixed throughout the remainder of the time step. The GS agent maintains a single speed and can choose from a continuous range of turn angles. The RL agent uses a discrete algorithm, deep Q-learning, and its action space contains five turn angles and two speeds. The RL agent can choose the same or half\footnote{Allowing the RL agent to decrease its speed is the small advantage the RL agent needs to win in fair initial conditions; otherwise, there are many situations in which the RL agent could never win against the GS agent.} the speed of the GS agent, and its choice of turn angles spans the same range as that of the GS agent.  The dynamics, $P$, of the environment are deterministic, as the state is fully observable and there are no stochastic effects such as wind. The targeting zone of an agent is a circular sector that emanates symmetrically from its heading; the radius is its targeting range, and the central angle, located at the agent's heading, is its targeting angle. The RL agent receives a reward, $R$, of 1 if the greedy shooter is within its targeting zone and receives a reward of 0 vice versa. If time expires or both agents are within each others' targeting zones, we count this as a loss for the RL agent and provide it with a reward of 0. We only supply terminal rewards and refrain from intermediate rewards, as these would unnecessarily influence the RL agent's actions. The discount factor, $\gamma$, dictates the extent to which the agent considers future rewards. Realized values of the action spaces and targeting zone\footnote{Due to ease of training, we choose unrealistically small values that can be scaled up to more appropriate magnitudes when desired.} are shown in Table \ref{vals}.

\begin{table}
	\caption{Action spaces and targeting zone.}
	\centering
	\begin{tabular}{lcl}
		\toprule
		Parameter & Value(s) & Units \\
		\cmidrule(r){1-3}
		RL Speed & $0.05,0.1$ & m\hspace{-0.1em}/\hspace{-0.1em}s \\
		RL Turn & $0, \pm\frac{\pi}{12}, \pm\frac{\pi}{6}$ & rad  \\
		\cmidrule(r){1-3}
		GS Speed & $0.1$ &  m\hspace{-0.1em}/\hspace{-0.1em}s \\
		GS Turn & $[-\frac{\pi}{6},\frac{\pi}{6}]$ & rad  \\
		\cmidrule(r){1-3}
		Targ. Range & $0.25$ & m \\
		Targ. Angle & $\frac{\pi}{6}$ & rad \\
		\bottomrule
		\label{vals}
	\end{tabular}
\end{table} 


\section{Methods}

\label{methods}

During training, the GS agent's position is initialized at the origin with heading in the positive \mbox{x-direction}. The RL agent's position is initialized randomly from a bivariate normal distribution centered at the origin with standard deviations of \mbox{0.5 \hspace{-0.1em}m} in the x- and y-directions, and the heading is chosen uniformly at random. Random initialization allows the agent some guaranteed wins after one time step; this provides free knowledge of victory states, considerably accelerating convergence. Episodes consist of 100 time steps, each one second long. A draw is declared if there is no winner after 100 time steps, which is identical to a loss for the RL agent in terms of rewards.

\clearpage

Although a 6-dimensional state is already quite small, we collapse the observed state down to three dimensions without losing any useful information through symmetries of the plane. Each observation translates and rotates the plane such that the greedy shooter is positioned at the origin with heading in the positive x-direction. This 3-dimensional vector details the relative location and heading of the RL agent; notably, this format is identical to the initial state of the environment. Exploiting symmetry decreases the state space and effectively increases the amount of data seen per unit of time.

To train the RL agent, we employ a deep Q-learning approach with a replay buffer \cite{2018Lapan, 2015Mnih, 2018Sutton}. The deep Q-network is fully connected and contains 4 hidden layers, each with 64 nodes. The input and output spaces are 3 and 10, respectively; each of the 5 possible angles and 2 speeds are paired to form 10 possible actions. A concern of performance divergence for the UAV pursuit-evasion problem with RL was discussed in \cite{2018Pippin}. We apply regularization through L2-norm weight decay in attempt to prevent this issue. Other notable hyperparameters are presented in Table \ref{hyper}. Our algorithm adopts the deep Q-learning code architecture from \cite{2018Lapan}, and our own code is on GitHub\footnote{\href{https://github.com/LorenJAnderson/uav-2d-greedyshooter-rl}{https://github.com/LorenJAnderson/uav-2d-greedyshooter-rl}}. 

Training win percentage provides a good summary of the RL agent's progress; however, the need for exploration mandates sub-optimal action selection during training, and as described previously, the initializations are random and often unbalanced. A better metric of success is win percentage of the RL agent without using any exploration using a standardized set of reasonably fair initializations. We provide a suite of 80 fixed test initializations where the agents start from 0.5\,m to 0.9\,m apart\footnote{The target range is 0.25 m which is at most half of the minimum start distance during testing.} with various headings. Training and testing results are produced simultaneously by testing the current RL agent's policy after every 25000 training time steps.

\begin{table}
	\caption{Algorithm hyperparameters.}
	\centering
	\begin{tabular}{ll}
		\toprule
		Hyperparameter & Value \\
		\cmidrule(r){1-2}
		Learning Rate & $1\mathrm{e}{-4}$ \\
		L2-norm weight decay & $1\mathrm{e}{-5}$ \\
		Discount Factor $\gamma$ & 0.99 \\
		Batch Size & 32\\
		Replay Buffer Size & $1\mathrm{e}{5}$\\
		Hidden Units / Layer & 64\\
		Hidden Layers & 4\\
		Final $\epsilon$-Greedy Probability & 0.02\\
		\bottomrule
		\label{hyper}
	\end{tabular}
\end{table}

\begin{figure}[h!]
	\includegraphics[trim={0cm 0cm 0cm 0cm}, clip, scale=0.52]{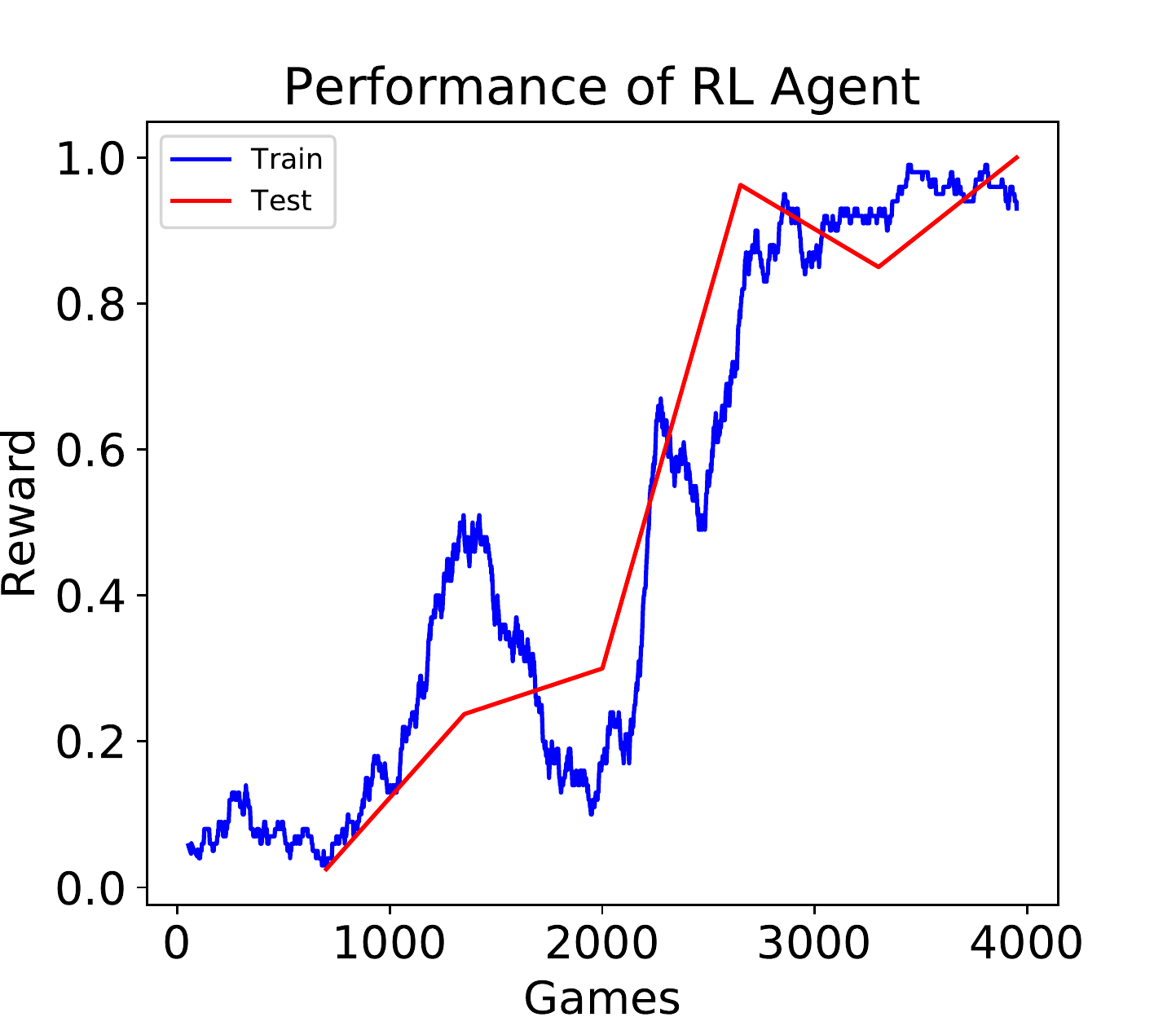}
	\caption{Training and testing dynamics of RL agent for one experiment. Rewards are terminal and therefore identical to win percentage. Training graph is smoothed by averaging the previous 100 rewards. }
	\label{traintest}
\end{figure}

\section{Results}

\label{result}

The training and testing results of one experiment are shown concurrently in Figure \ref{traintest}. Testing occurred 6 times; the last three policies won the majority of games, and the final policy, which we analyze throughout the remainder of this section, had perfect testing performance. We ran each experiment until the RL agent achieved perfect testing performance, the duration averaging 10 to 15 minutes. We did not see any performance divergence, as the training curve would rarely fall below 0.8 even hours after achieving a perfect testing score. 

\clearpage

\begin{figure}[h!]
	\includegraphics[trim={0cm 0cm 0cm 0cm}, clip, scale=0.48]{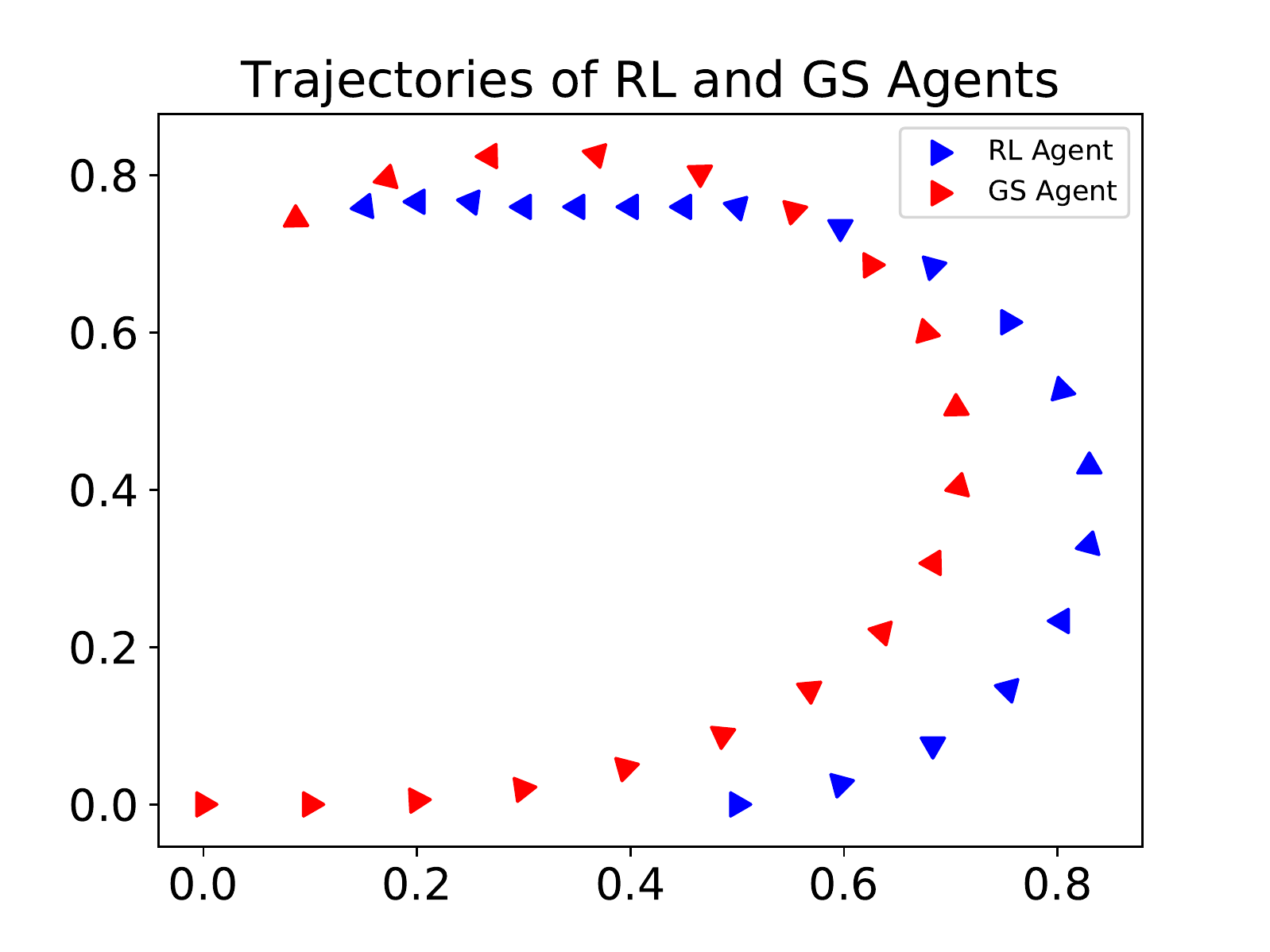}
	\caption{One sample set of trajectories from the test suite. The RL agent moved in front of the GS agent and decreased speed to win the engagement.}
	\label{sampletraj}
\end{figure}

We now showcase the RL agent's prowess by highlighting sample trajectories: sequences of states, actions, and rewards. Figure \ref{sampletraj} displays the most common tactic that the RL agent employs to win. Here, the RL agent turns in front of the GS agent, carefully avoiding its targeting zone, and immediately decreases speed. The GS agent then turns in the direction of the RL agent, moving at double the speed. Eventually the GS agent is positioned ahead of the RL agent and is caught in its targeting zone. This tactic provides the RL agent flexibility to increase its speed to maintain close proximity to the GS agent if necessary.

\begin{figure}[h!]
	\includegraphics[trim={0cm 0cm 0cm 0cm}, scale=0.48]{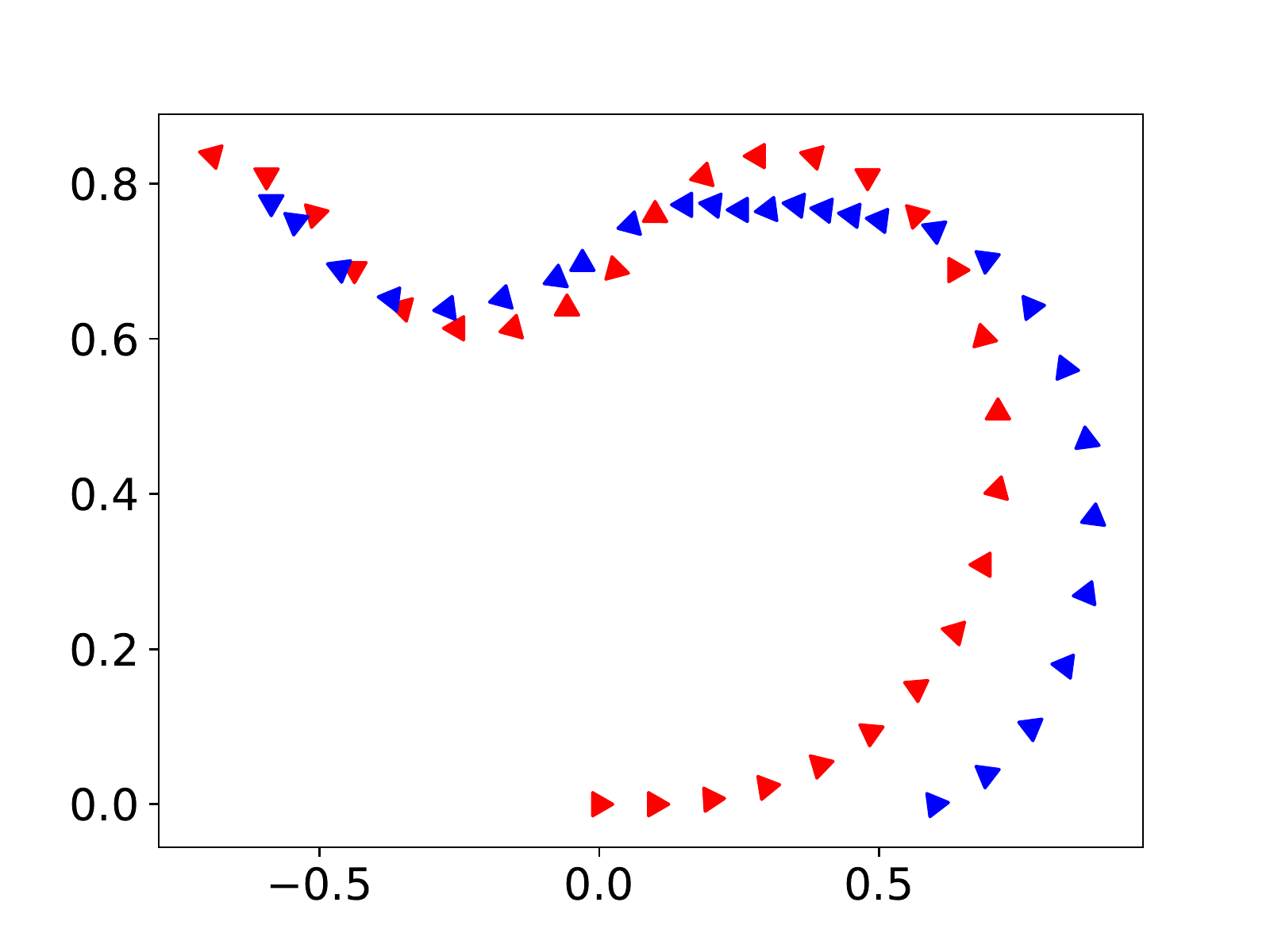}
	\caption{}
	\label{2p}
\end{figure}

\begin{figure}[h!]
	\includegraphics[trim={0cm 0cm 0cm 0cm}, scale=0.48]{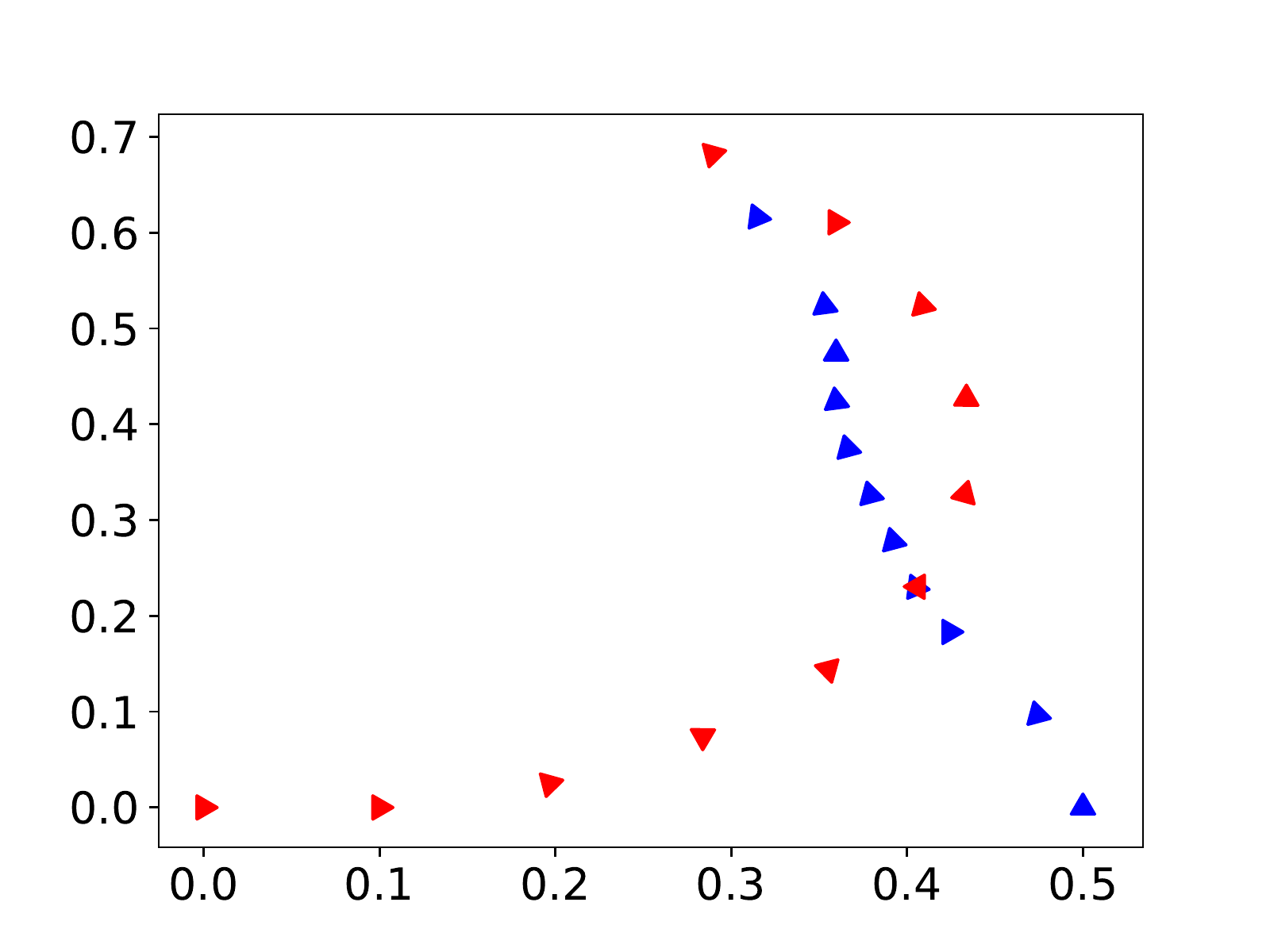}
	\caption{}
	\label{11p}
\end{figure}

\begin{figure}[h!]
	\includegraphics[trim={0cm 0cm 0cm 0cm}, scale=0.48]{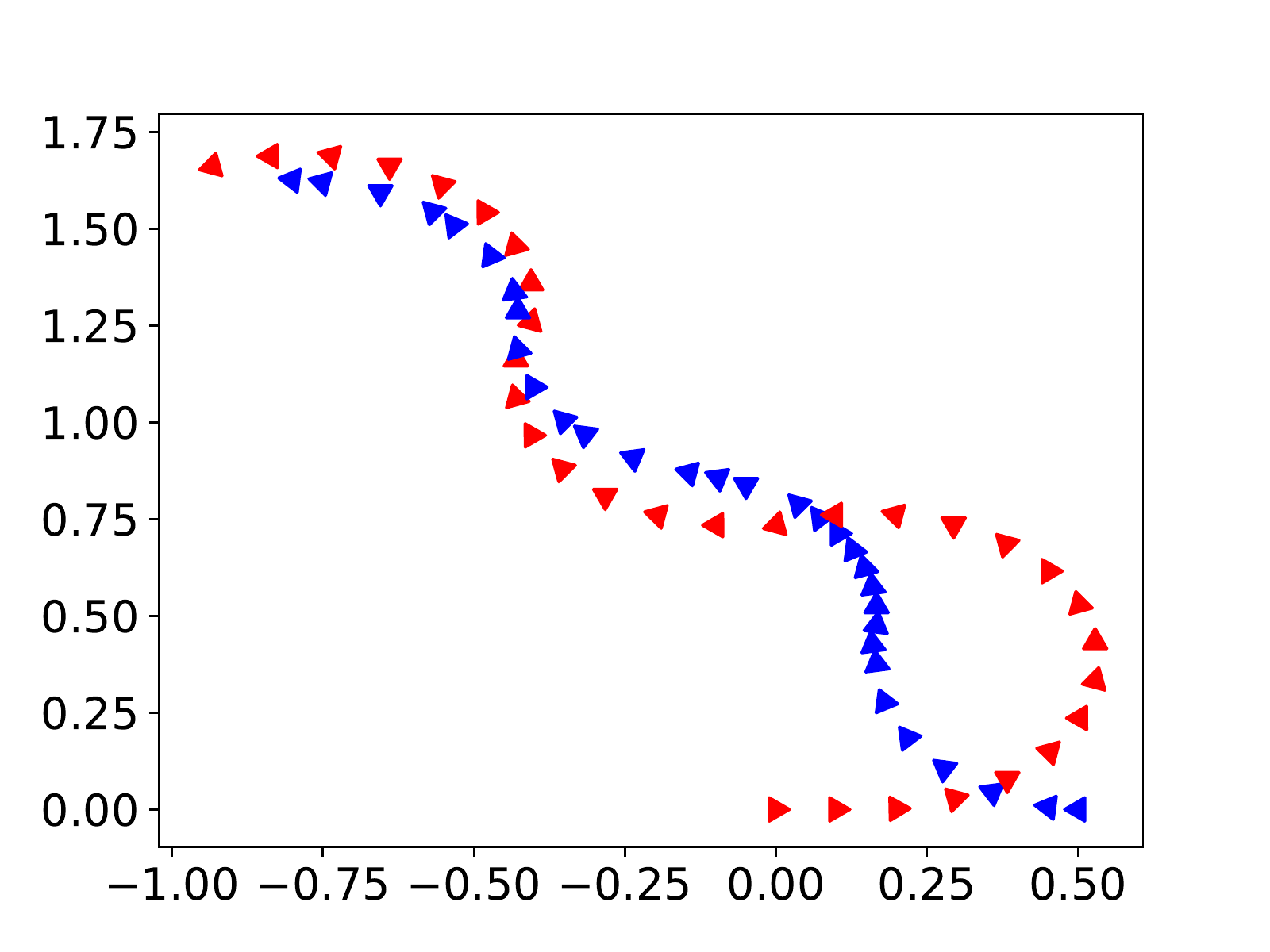}
	\caption{}
	\label{21p}
\end{figure}

\begin{figure}[h!]
	\includegraphics[trim={0cm 0cm 0cm 0cm}, scale=0.48]{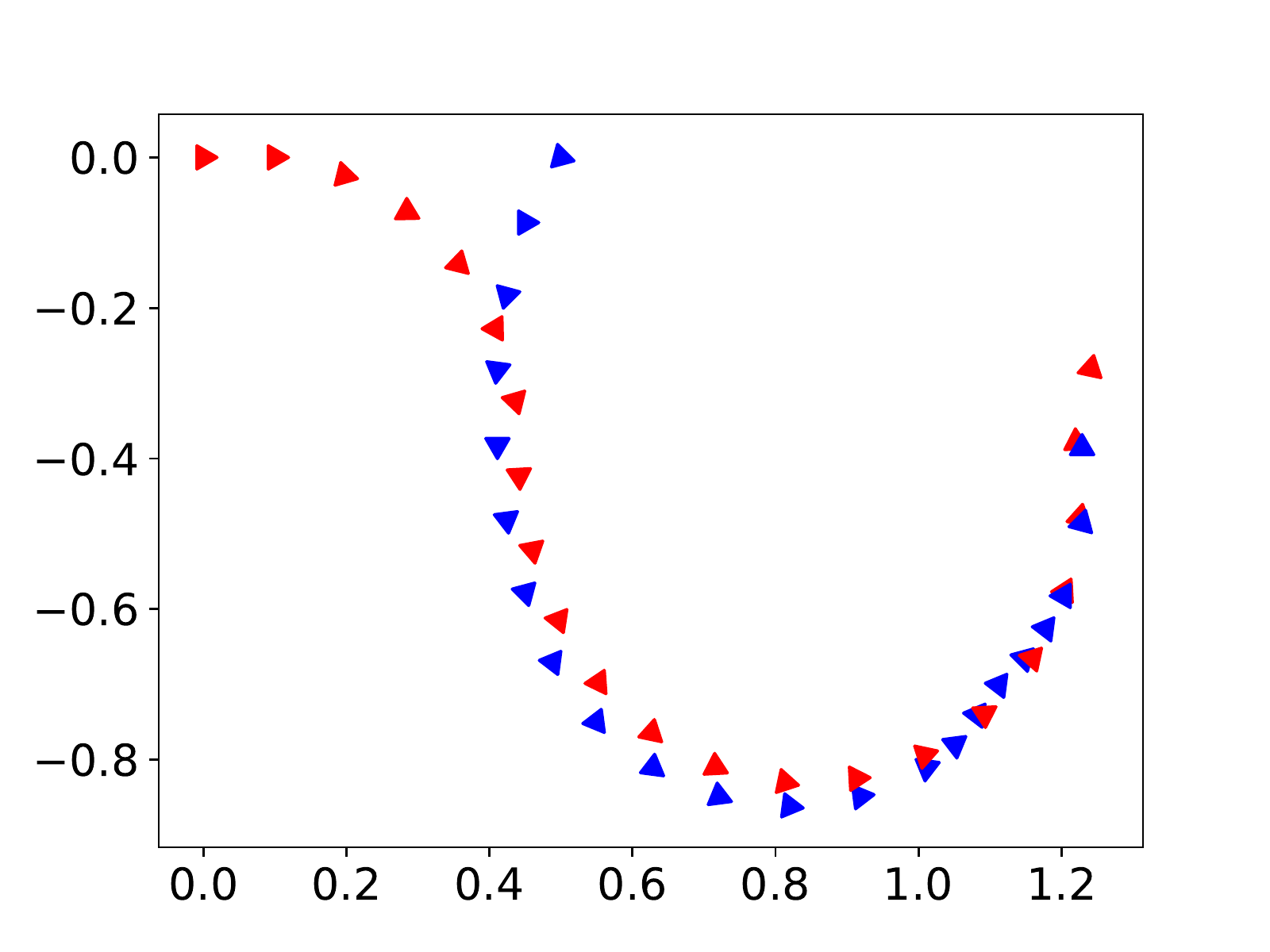}
	\caption{}
	\label{26p}
\end{figure}

\begin{figure}[h!]
	\includegraphics[trim={0cm 0cm 0cm 0cm}, scale=0.48]{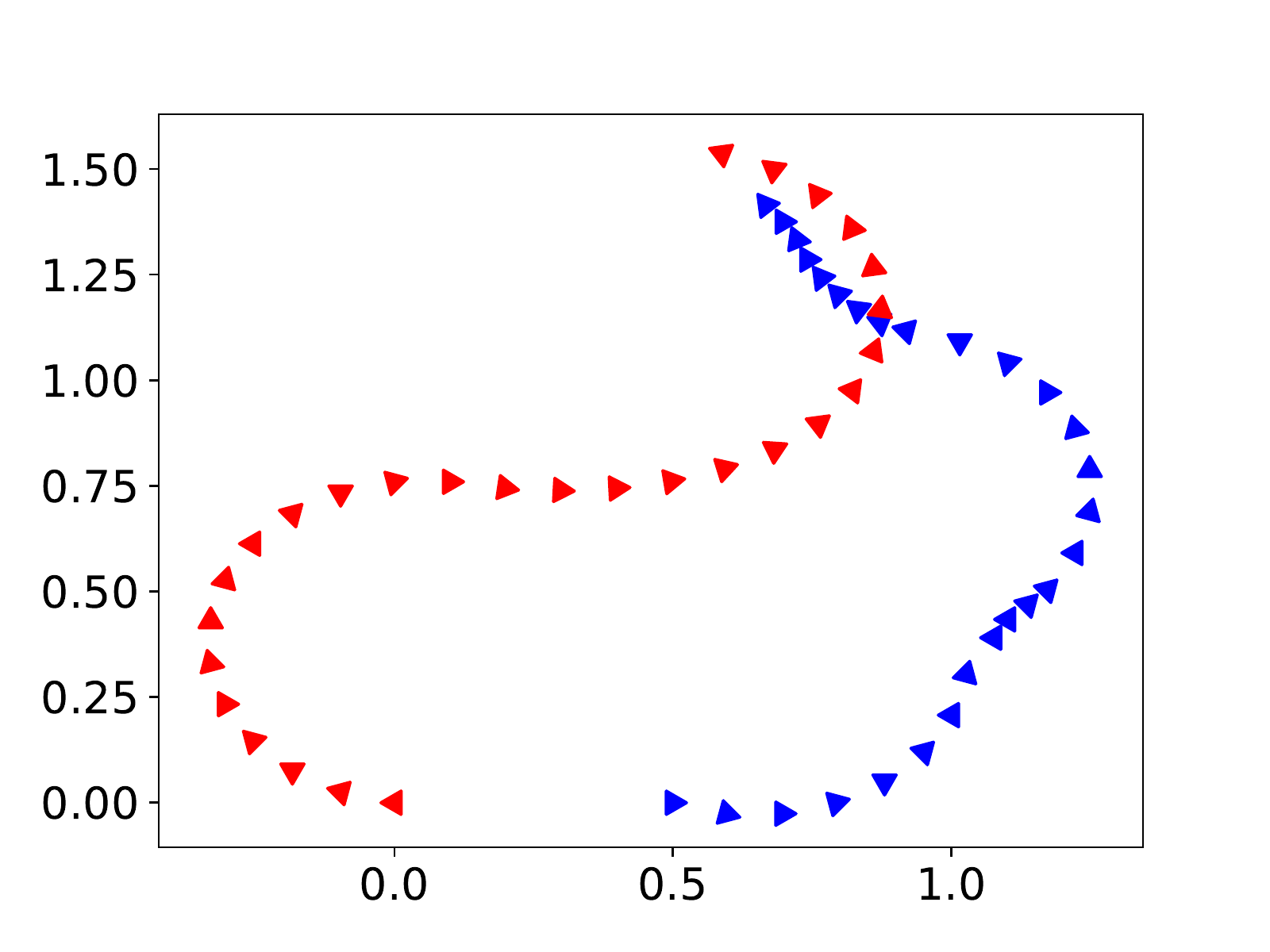}
	\caption{}
	\label{41p}
\end{figure}

\begin{figure}[h!]
	\includegraphics[trim={0cm 0cm 0cm 0cm}, scale=0.48]{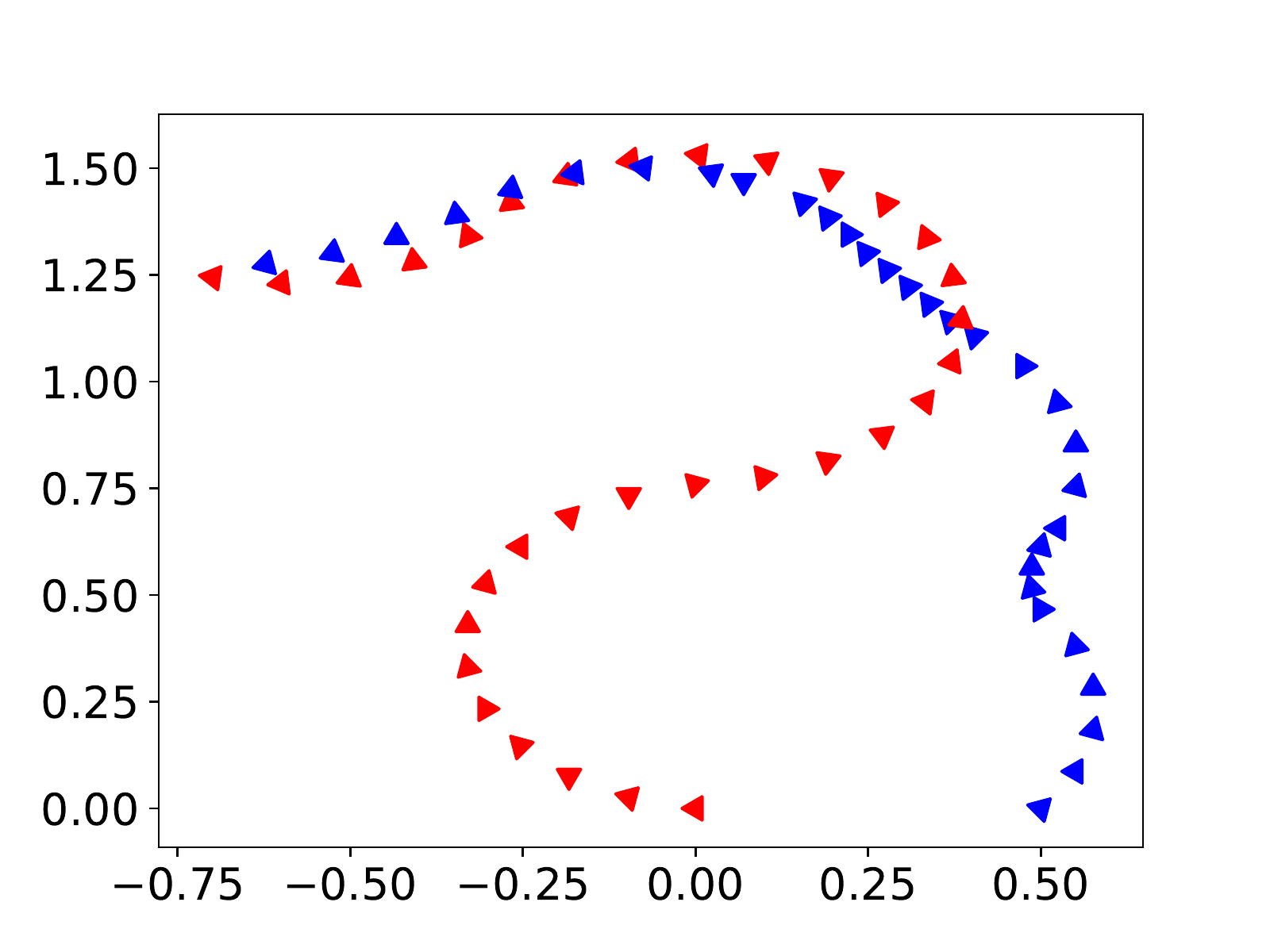}
	\caption{}
	\label{46p}
\end{figure}

\begin{figure}[h!]
	\includegraphics[trim={0cm 0cm 0cm 0cm}, scale=0.48]{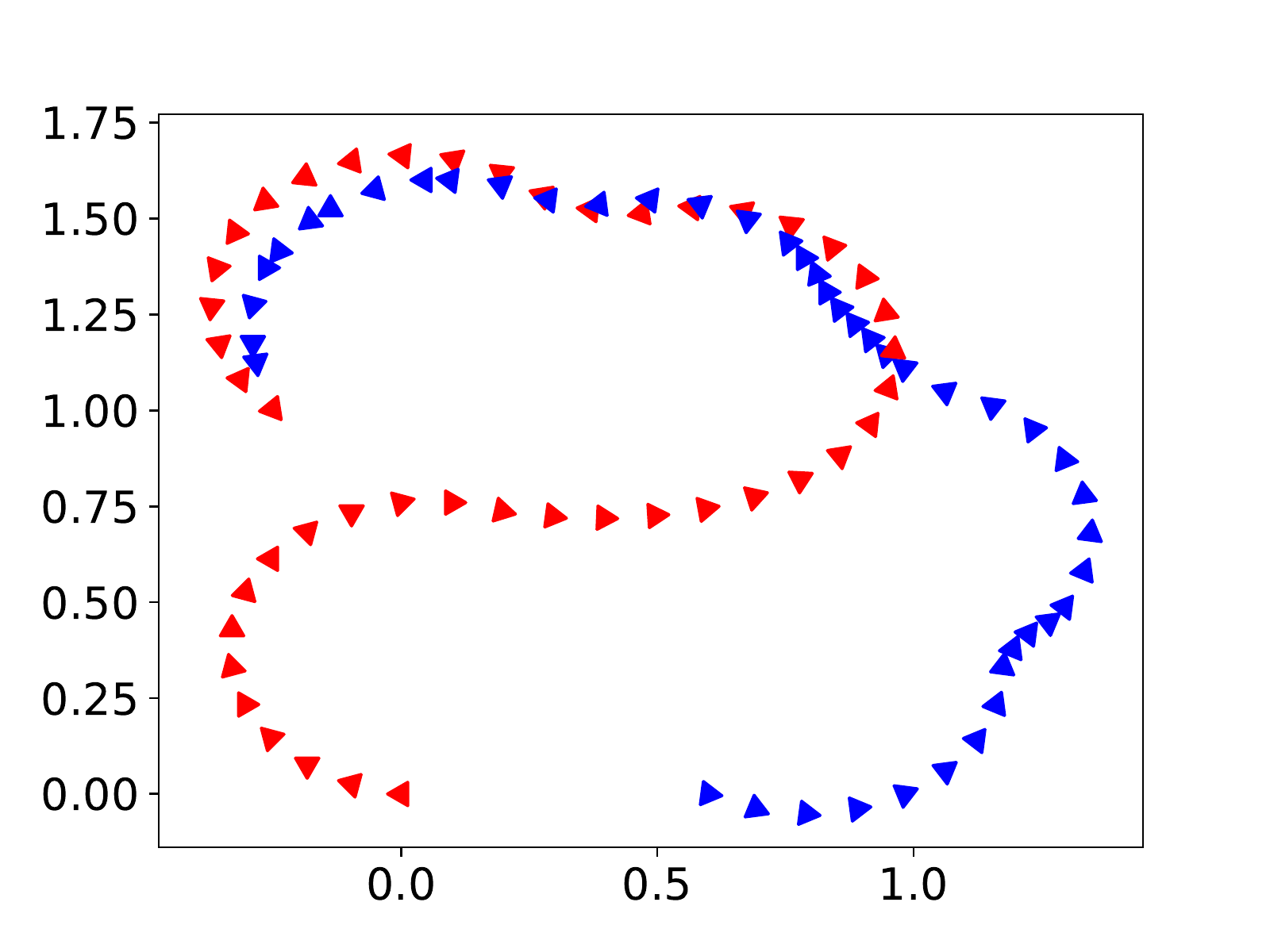}
	\caption{}
	\label{47p}
\end{figure}

\begin{figure}[h!]
	\includegraphics[trim={0cm 0cm 0cm 0cm}, scale=0.48]{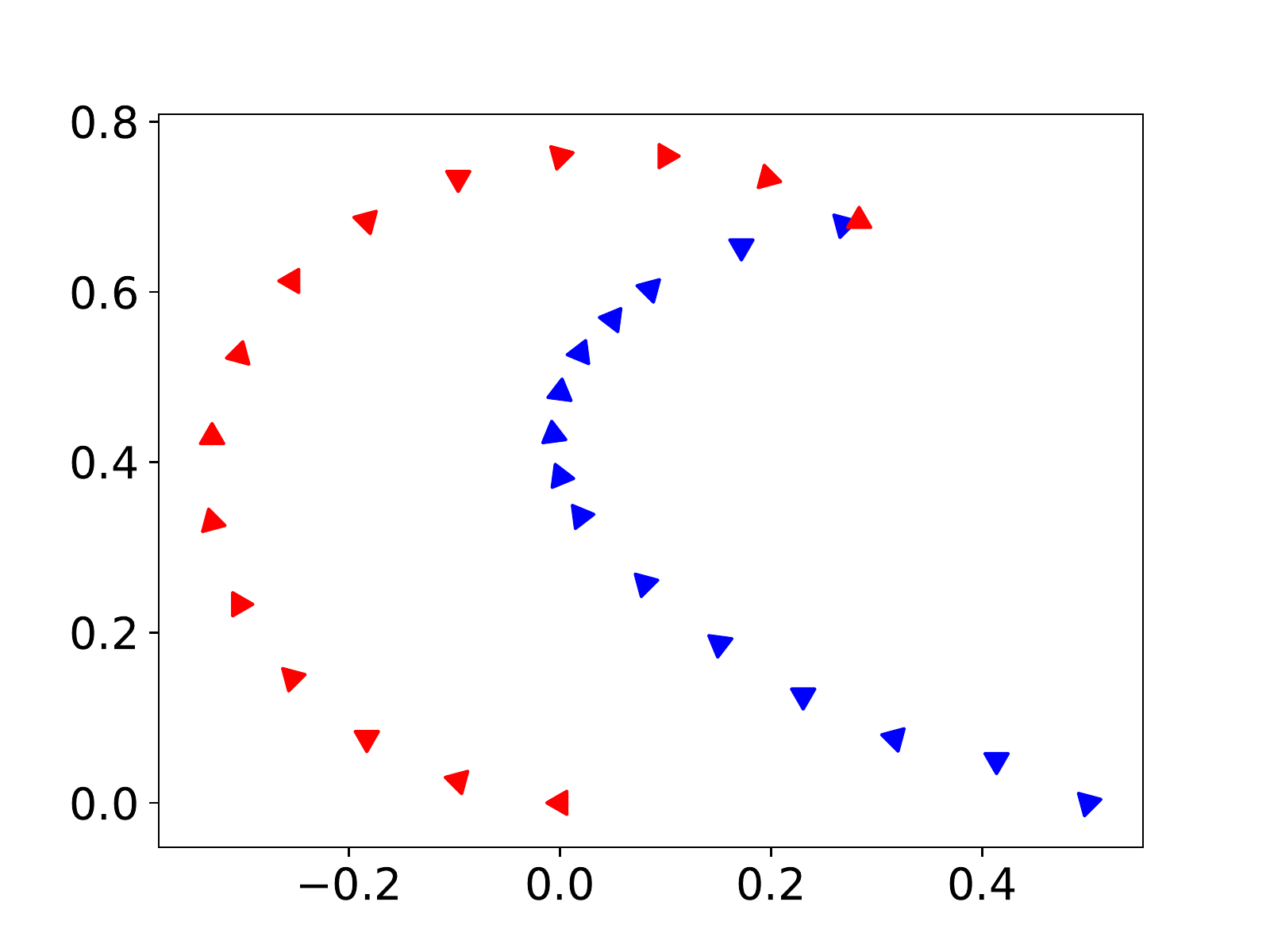}
	\caption{}
	\label{56p}
\end{figure}

\begin{figure}[h!]
	\includegraphics[trim={0cm 0cm 0cm 0cm}, scale=0.48]{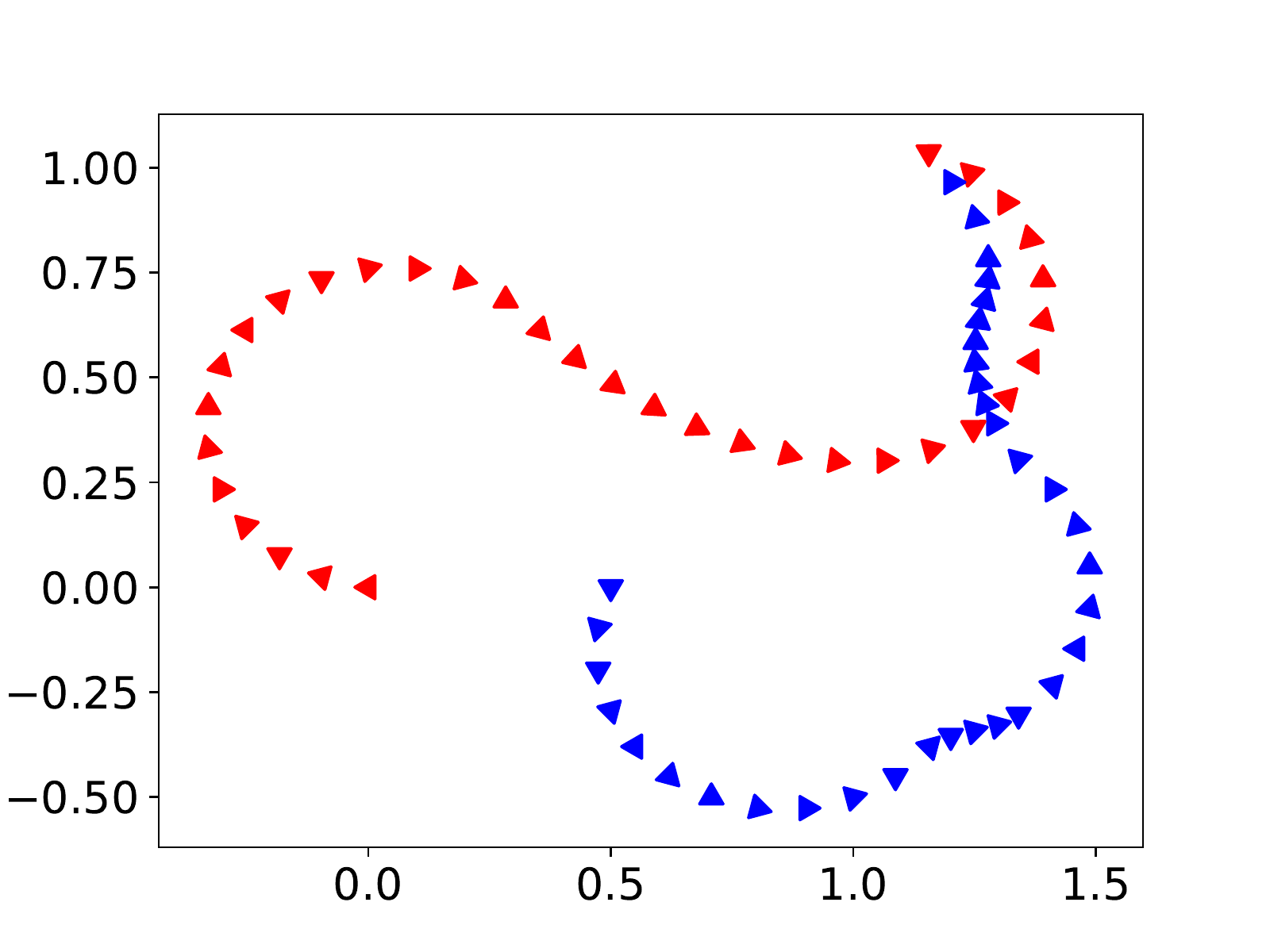}
	\caption{}
	\label{71p}
\end{figure}

Next, we discuss trajectories found in the appendix. Figures \ref{25p} and \ref{51p} exhibit the most common trajectories, similar to that in Figure \ref{sampletraj}. Figures \ref{2p} and \ref{21p} show how the RL agent acts in serpentine fashion when behind the GS agent to continually decrease their separation. The RL agent is not restricted to cutting in front of the GS agent as Figures \ref{26p} and \ref{56p} display. Also, Figures \ref{11p} and \ref{56p} show that the agent is opportunistic and is aware of short-term capture possibilities. In Figures \ref{41p}, \ref{46p}, \ref{47p}, and \ref{71p}, the agent makes notably long-term decisions by varying both speed and turn angle early in the trajectory. In contrast to \cite{2018Pippin}, we see structured as opposed to random actions when the two agents are distant from each other.


\section{Conclusions \& Future Work}

\label{con}

We trained an RL agent to successfully solve the 2D greedy shooter problem for UAVs in simulation. This method provides free knowledge of victory states through random initialization, exploits planar symmetries to reduce the state space, and incorporates weight decay to prevent performance divergence. The learned policies are opportunistic and farsighted, demonstrating perfect performance on a test suite of 80 various initializations.

Since our work was performed entirely in simulation, we first plan to pursue more realistic and complex environments such as SCRIMMAGE \cite{2019DeMarco} or  AirSim \cite{2018Shah}. Pending successful results, we will test our policies in physical domains. Another possibility is to extend our work to 3D environments, as they have been explored much less frequently than their 2D counterparts. Other ideas involve restricting action choice by adding penalties through fuel costs or presenting noisy observations through partial or competitive observability, such as in \cite{2012Li}. Our long-term goal is to extend this work to the emerging field of multi-agent RL.


\clearpage

\section{Appendix}
\label{app}

We display 11 distinct trajectories of the RL and GS agents from the test suite. The RL agent is in blue and the GS agent is in red. Each figure illustrates one game, and each marker represents one time step. All games initialize the GS agent at the origin. Initial headings are varied. The RL agent won every engagement by catching the GS agent in its targeting zone before 100 time steps elapsed. Note how the RL agent wins by reducing speed in key locations.
\vspace{2.75em}

\vspace{-1.4em}

\begin{figure}[h!]
	\includegraphics[trim={0cm 0cm 0cm 0cm}, scale=0.48]{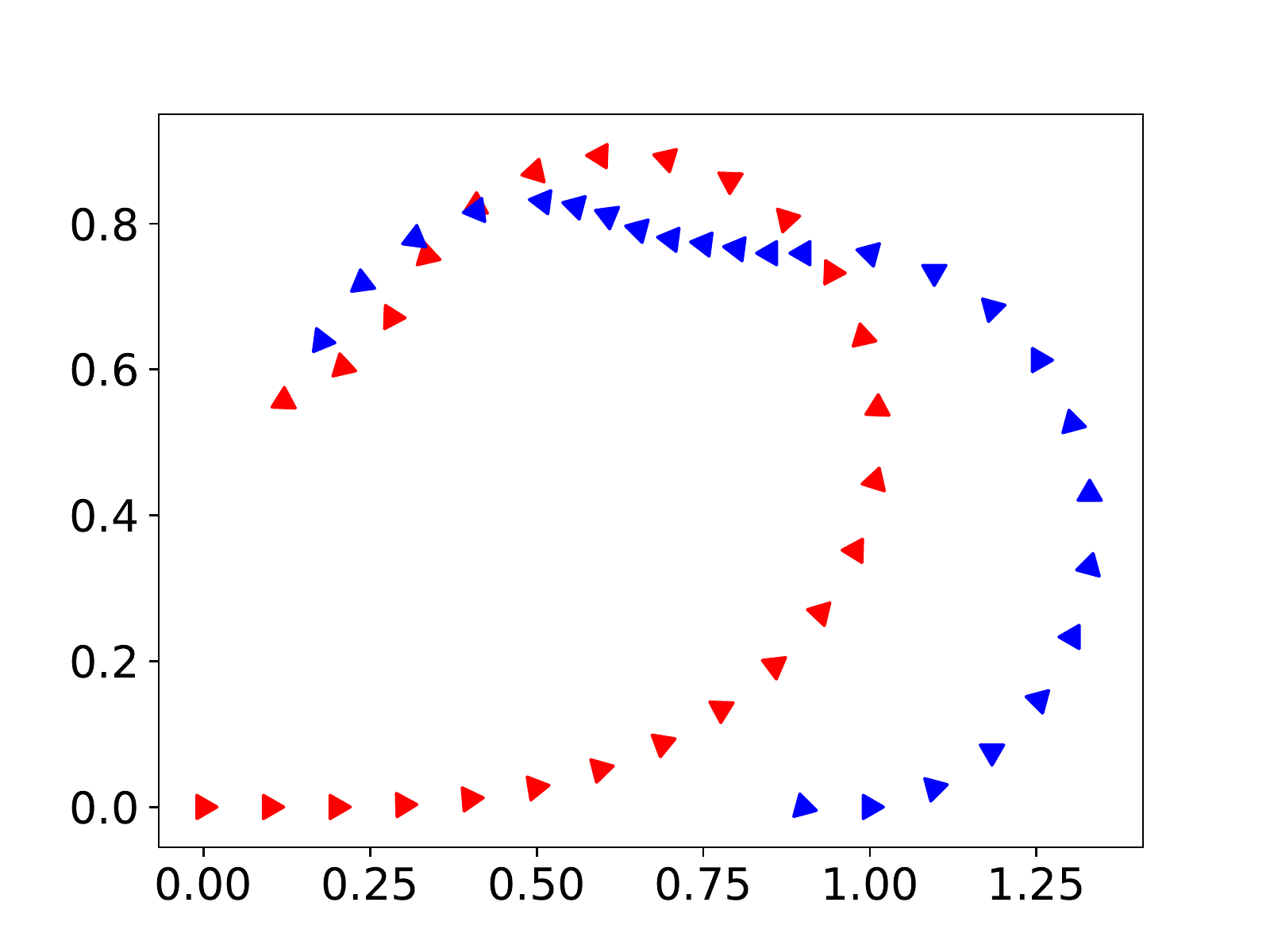}
	\caption{}
	\label{25p}
\end{figure}

\begin{figure}[h!]
	\includegraphics[trim={0cm 0cm 0cm 0cm}, scale=0.48]{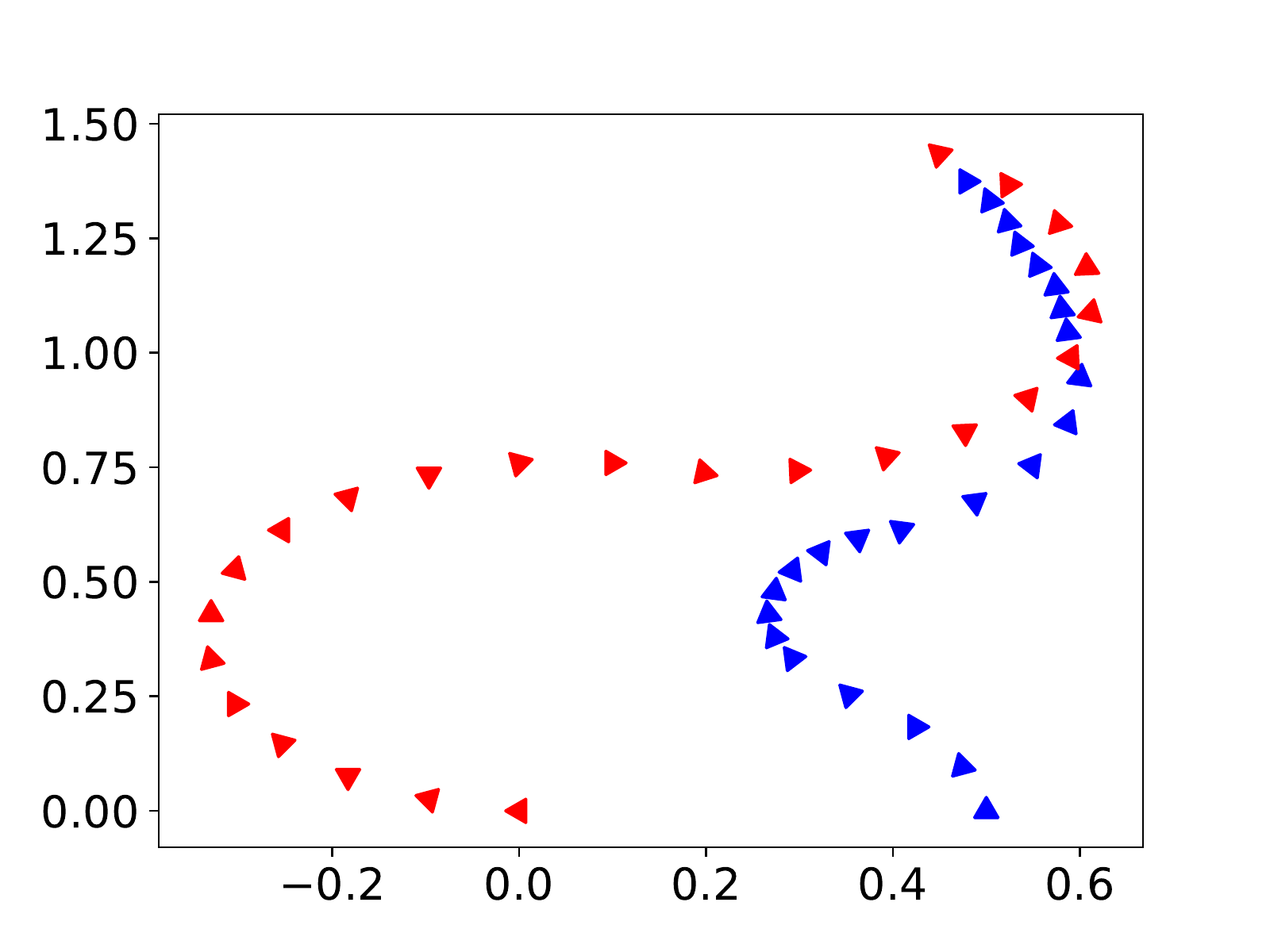}
	\caption{}
	\label{51p}
\end{figure}

\end{document}